\documentclass[letterpaper, 10 pt, conference]{ieeeconf}  

\IEEEoverridecommandlockouts                              

\usepackage{algorithmic}
\usepackage{array}
\usepackage[caption=false,font=normalsize,labelfont=sf,textfont=sf]{subfig}
\usepackage{siunitx}
\usepackage{textcomp}
\usepackage{stfloats}
\usepackage{url}
\usepackage{verbatim}
\usepackage{graphicx}

\usepackage{graphics} 
\usepackage{epsfig} 
\usepackage{mathptmx} 
\usepackage{times} 
\usepackage{amsmath} 
\usepackage{amssymb}  

\usepackage[hidelinks]{hyperref}

\title{\LARGE \bf
Rhino: An Autonomous Robot for Mapping Underground Mine Environments
}

\newcommand{\orcidicon}{\includegraphics[width=0.32cm]{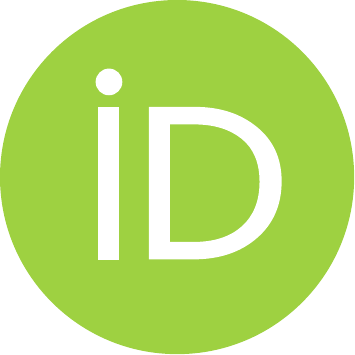}}
\expandafter\xdef\csname orcidCT\endcsname{\noexpand\href{https://orcid.org/\csname orcidauthorCT\endcsname}{\noexpand\orcidicon}}
\expandafter\xdef\csname orcidDC\endcsname{\noexpand\href{https://orcid.org/\csname orcidauthorDC\endcsname}{\noexpand\orcidicon}}
\expandafter\xdef\csname orcidJB\endcsname{\noexpand\href{https://orcid.org/\csname orcidauthorJB\endcsname}{\noexpand\orcidicon}}
\expandafter\xdef\csname orcidBT\endcsname{\noexpand\href{https://orcid.org/\csname orcidauthorBT\endcsname}{\noexpand\orcidicon}}
\expandafter\xdef\csname orcidYG\endcsname{\noexpand\href{https://orcid.org/\csname orcidauthorYG\endcsname}{\noexpand\orcidicon}}

\author{Christopher Tatsch$^{1}$\orcidCT{}, Jonas Amoama Bredu Jnr$^{1}$\orcidJB{}, Dylan Covell$^{1}$\orcidDC,  Ihsan Berk Tulu$^{2}$\orcidBT{}, Yu Gu$^{1}$\orcidYG{}
\thanks{*This work was not supported by any organization}
\thanks{$^{1}$ are with the Department of Mechanical and Aerospace Engineering, West Virginia University, Morgantown, WV 26505, USA
     {\tt\small  christophertatsch@ieee.org; yugu@mail.wvu.edu)}
}
\thanks{$^{2}$ is with the Department of Mining Engineering, West Virginia University, Morgantown, WV 26505, USA 
 }%
}

\begin{document}

\maketitle
\thispagestyle{empty}
\pagestyle{empty}

\begin{abstract}
There are many benefits for exploring and exploiting underground mines, but there are also significant risks and challenges. One such risk is the potential for accidents caused by the collapse of the pillars, and roofs which can be mitigated through inspections. However, these inspections can be costly and may put the safety of the inspectors at risk. To address this issue, this work presents Rhino, an autonomous robot that can navigate underground mine environments and generate 3D maps. These generated maps will allow mine workers to proactively respond to potential hazards and prevent accidents. The system being developed is a skid-steer, four-wheeled unmanned ground vehicle (UGV) that uses a LiDAR and IMU to perform long-duration autonomous navigation and generation of maps through a LIO-SAM framework. The system has been tested in different environments and terrains to ensure its robustness and ability to operate for extended periods of time while also generating 3D maps.
\end{abstract}


\section{Introduction}

\noindent Mining operations are required to extract necessary minerals in nearly every industry. Minerals such as copper, zinc, and others are necessary for electric and hybrid vehicles, and silver and copper are useful for medical applications because of their inherent antimicrobial properties \cite{mine_future}. Traditionally, surface mining methods have been the major producer of minerals, however, for deeper orebodies, surface mining becomes uneconomical, and underground mining methods are applied to extract valuable minerals. However, there are many risks involved in the exploration and exploitation of these environments. For example, there were $98$ fatalities and $14,697$ injuries in the US underground mining industry between 2012 to 2022 \cite{mine_fatality_report}. 
Proper inspection to identify hazards and deteriorating ground conditions and rapid response to these hazardous conditions would protect miners
from possible accidents \cite{esterhuizen2011pillar}.
Due to the size of some underground mines, the inspections are economically infeasible and dangerous for humans to complete. 
LiDAR-based scans can provide better and more detailed data for structural analysis of the mine, however, they are constrained to a fixed face of a pillar and equipment requires manual operation \cite{slaker2015monitoring}. 
Robots can greatly contribute to improving mining safety. One way is that it allows remote inspection, not requiring a human to be in a dangerous environment during the process. It also enables the  structural inspection of the mine by creating large high-fidelity maps of the mine, and by using a mobile platform that can autonomously perform localization and mapping. This automation, without direct human supervision, would allow for more frequent and detailed inspection of the environment to improve safety.

\begin{figure}[t]
\centering
\includegraphics[width=0.90\linewidth]{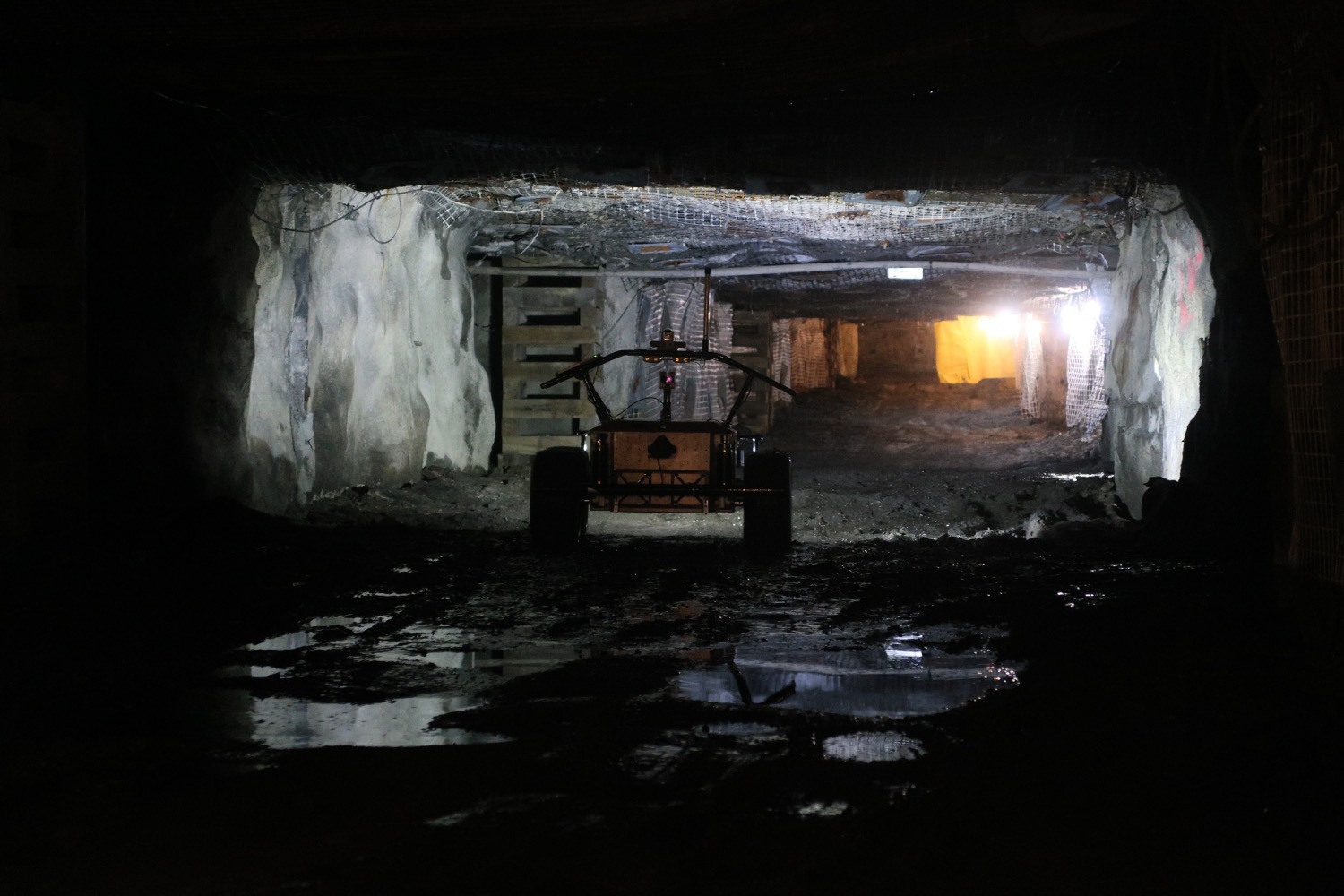}
\caption{Rhino robot inside an experimental mine during a mapping task \label{fig:rhino_mine}}
\end{figure}   
\unskip

Operating autonomous robots in subterranean spaces is very challenging due to the harsh environments. These are very large environments that can cover several square kilometers, and also with large structures to inspect, such as in stone mines with pillar widths of up to $\SI{21.5}{\meter}$ and heights of up to $\SI{38}{\meter}$  \cite{esterhuizen2011pillar}. The underground mines are also GPS-denied environments where the robot cannot rely on satellite-based navigation for updating its state estimation during operation. The terrain is difficult to traverse, with slippery surfaces, large rocks, and mud. Obscurants in these environments such as dust, smoke, and fog degrade the performance sensors and may cause inaccurate measurements. Communication is very limited, and only reliable with direct line-of-sight due to the thickness of the walls \cite{yarkan2009underground}. It is also dangerous for humans to conduct these inspections, therefore, the robot needs to be capable of autonomously traversing through waypoints while choosing a safe path to follow and avoiding obstacles for mapping the roofs and pillars in the mines.

There have been many advances in robot technologies for underground mapping and exploration with the DARPA Subterranean Challenge \cite{darpa_challenge} that enabled autonomous robot operation in these environments. The competition aimed to improve search and rescue efforts in these environments, where time and knowledge of environmental hazards are critical to finding survivors. The implementation of robotic systems allows for the rapid creation of maps, and the localization of artifacts can greatly help survivors and rescue teams.
The developments in software systems include the improvement of autonomy algorithms like NeBuLa \cite{agha2021nebula}, communication systems between multi-robot systems \cite{rouvcek2021system}, traversability of subterranean environments \cite{fankhauser2014robot}, and significant advancements in SLAM for underground and GPS-denied environments \cite{ebadi2022present}.
LiDAR sensors are widely used in underground environments for localization and mapping due to their versatility and ability to work in various environments and conditions. The registration methods for these environments include methods that extract features between consecutive frames such as \cite{khattak2020complementary} and \cite{zhao2021super}, or by methods that use dense registration of LiDAR points using Iterative Closest Point (ICP) and its variants \cite{ramezani2022wildcat}, \cite{kubelka2022gravity}, \cite{reinke2022locus}. Loop closure candidates are chosen using methods such as bag-of-visual-words \cite{galvez2012bags} and distance-based loop closures \cite{xue2022lego}. Inter robot loop closure is explored in DOOR-SLAM \cite{lajoie2020door}. The standard SLAM back-end solution for these algorithms is a non-linear maximum a-posteriori estimation via factograph, that is implemented with solvers such as \cite{dellaert2012factor}. Solutions that were developed for underground mines include CompSLAM \cite{khattak2020complementary}, Wildcat SLAM \cite{ramezani2022wildcat}, LOCUS/LAMP \cite{agha2021nebula} and LIO-SAM \cite{shan2020lio}.

The DARPA Challenge showcased the collaboration of a variety of systems, some of which were adaptations of commercial products, as well as custom platforms for unmanned ground and aerial vehicles with operational times of an hour for UGVs and $20$ minutes for UAVs. There was a variety of legged (walking and wheeled), hybrid (ground/aerial), tracked, and aerial robots \cite{rouvcek2020darpa,tranzatto2022cerberus,hudson2021heterogeneous}. 
These solutions, however, are designed to operate in harsh environments for short periods of time. The competition allocated a period of $1$ hour for each team to complete all the tasks. However, the limitation of the operational time might not be suitable for the mapping and inspections of large underground mines.


The research team developed Rhino, shown in Figure \ref{fig:rhino_mine} for the inspection of large underground stone mine environments. Rhino is a four-wheeled, split-body robot design developed for long-term autonomous operations in underground mines. Rhino is also designed to carry a tethered drone, its winch system, and batteries as a payload, to extend the aerial vehicle's operational time and to inspect difficult-to-reach areas \cite{martinez2023oxpecker}. This work's contributions include the development of a new robot design suitable for the rough underground environment previously described, improving the algorithm to support long-term autonomous operation, and field testing in the mine environment. 

The rest of this paper is organized as follows: Section \ref{sec:robot_design} describes the mechanical and electrical design of the robot. Section \ref{sec:robot_software_and_operations} describes the software architecture of the robot and how it is designed to operate in the mine environment. Section \ref{sec:experiments_and_results} describes the tests that were performed to validate the robot's capabilities for operating in underground mines, demonstrating localization, planning, obstacle avoidance capabilities, and generating high-fidelity maps. The conclusion and future work are presented in Section \ref{sec:conclusion} 


\section{Robot Design}
\label{sec:robot_design}

Since the Rhino platform is being developed to assist in the monitoring of underground mines, the objectives of the UGV are \cite{jonasthesis}:
\begin{enumerate}
    \item Traverse the mine terrain reliably and autonomously;
    \item Avoid obstacles and untraversable areas;
    \item Create high-resolution 3D maps of the mine and columns;
    \item Operate for extended periods of time of at least six hours;
    \item Support the tethered drone system.
\end{enumerate}


\subsection{Mechanical Overview}

Due to the limitations in the current energy density of batteries (e.g., around $\SI{295}{\watt\hour\per\kilogram}$ for Lithium-Ion \cite{li2017toward}), the weight of the battery was the main constraint when designing a robot that is able to operate continuously for six hours. To successfully complete the required tasks in the mine, a wheeled design was chosen due to its efficiency with four large wheels to allow for the traversal of the rugged terrain. A skid-steering driving configuration is used due to its simplicity, and ease of service. 
A split-body chassis was chosen because it allows all four $\SI{0.5}{\meter}$ wheels to come into contact with the ground to provide traction on uneven terrain. A $\SI{300}{\mm}$ diameter bearing connects both halves instrumented along the body axis of the robot. 

\begin{figure}[b]
\centering
\includegraphics[width=0.90\linewidth]{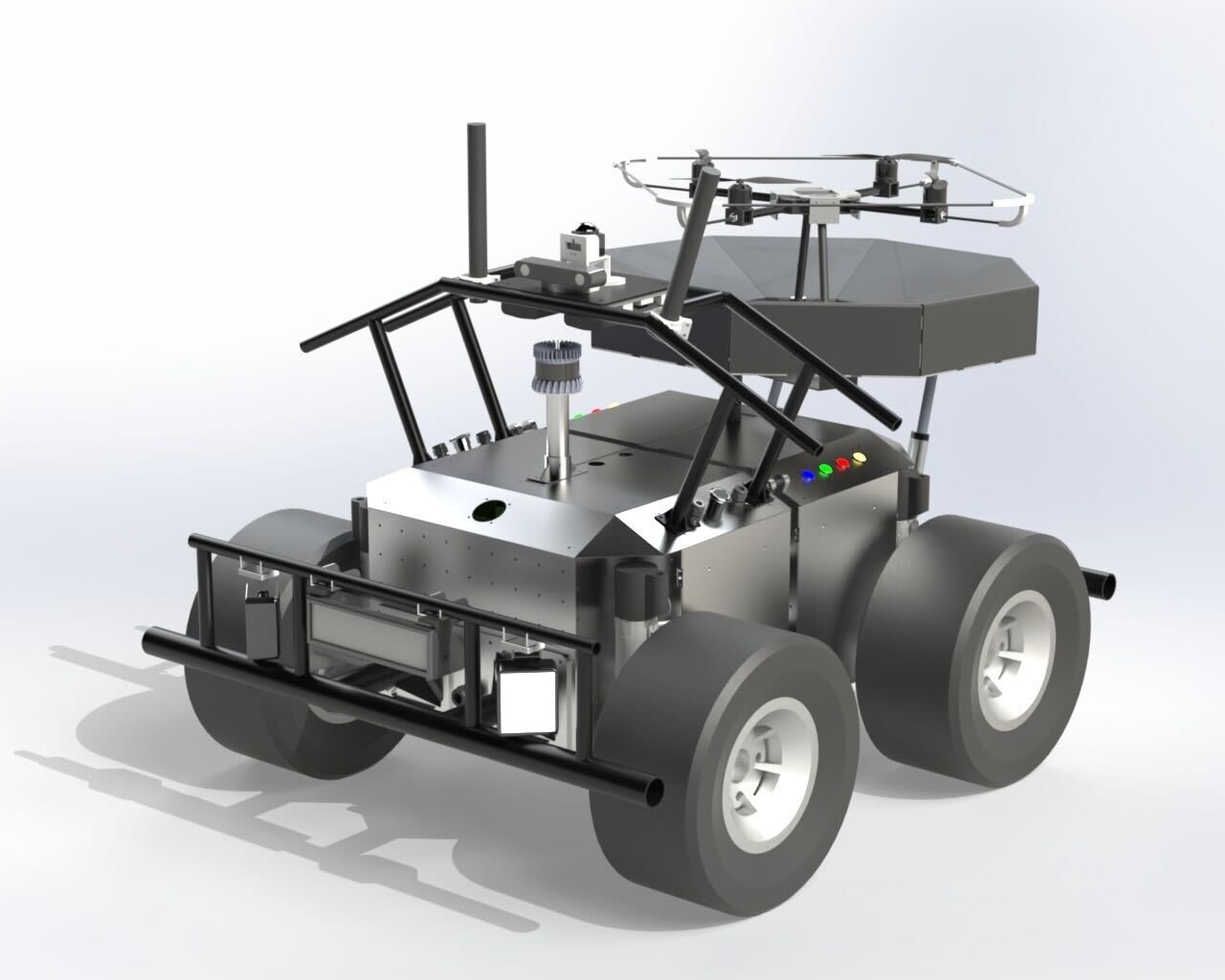}
\caption{Rhino robot rendering, including sensors, landing platform, and the drone \label{fig:rhino_rendering}}
\end{figure}   
\unskip

Lithium Iron Phosphate (LiFePO4) batteries were chosen due its longer life span and safety. Ten 12.8V batteries with a combined energy of $\SI{420}{\A\hour}$ and a combined weight of $\SI{58}{\kilogram}$ are used to power the electronics and sensors. A render of the robot is shown in Figure \ref{fig:rhino_rendering}. The robot is divided into two parts: the front chassis and the back chassis. There are six batteries in the back chassis of the robot, with an easy-to-access front panel with a connection to charge the batteries. The front chassis has two side compartments with one side instrumented with electronics required to power the system, while the other compartment contains the robot's computer. In the back chassis, there are six batteries and the winch system for the drone. The drone and its landing platform can be mounted on top of the back chassis. The two side compartments in the back are instrumented with the motor controllers and other auxiliary electronics such relays, network switch, etc. All the panels and compartments are sealed for insulation for operating inside the mine environment. A roll cage is mounted on the top of the front chassis of the rover for protecting the chassis as well as for mounting cameras and communication antennas and radios. Front and back bumpers are added for additional safety and for mounting the lights. The robot's total weight is $\SI{190}{\kilogram}$ and its footprint is $\SI{1.52}{\meter}$ by $\SI{1.15}{\meter}$ by $\SI{1.15}{\meter}$ and has a ground clearance of $\SI{0.22}{\meter}$.

\subsection{Electronics Overview}

\begin{figure}
\centering
\includegraphics[width=0.95\linewidth]{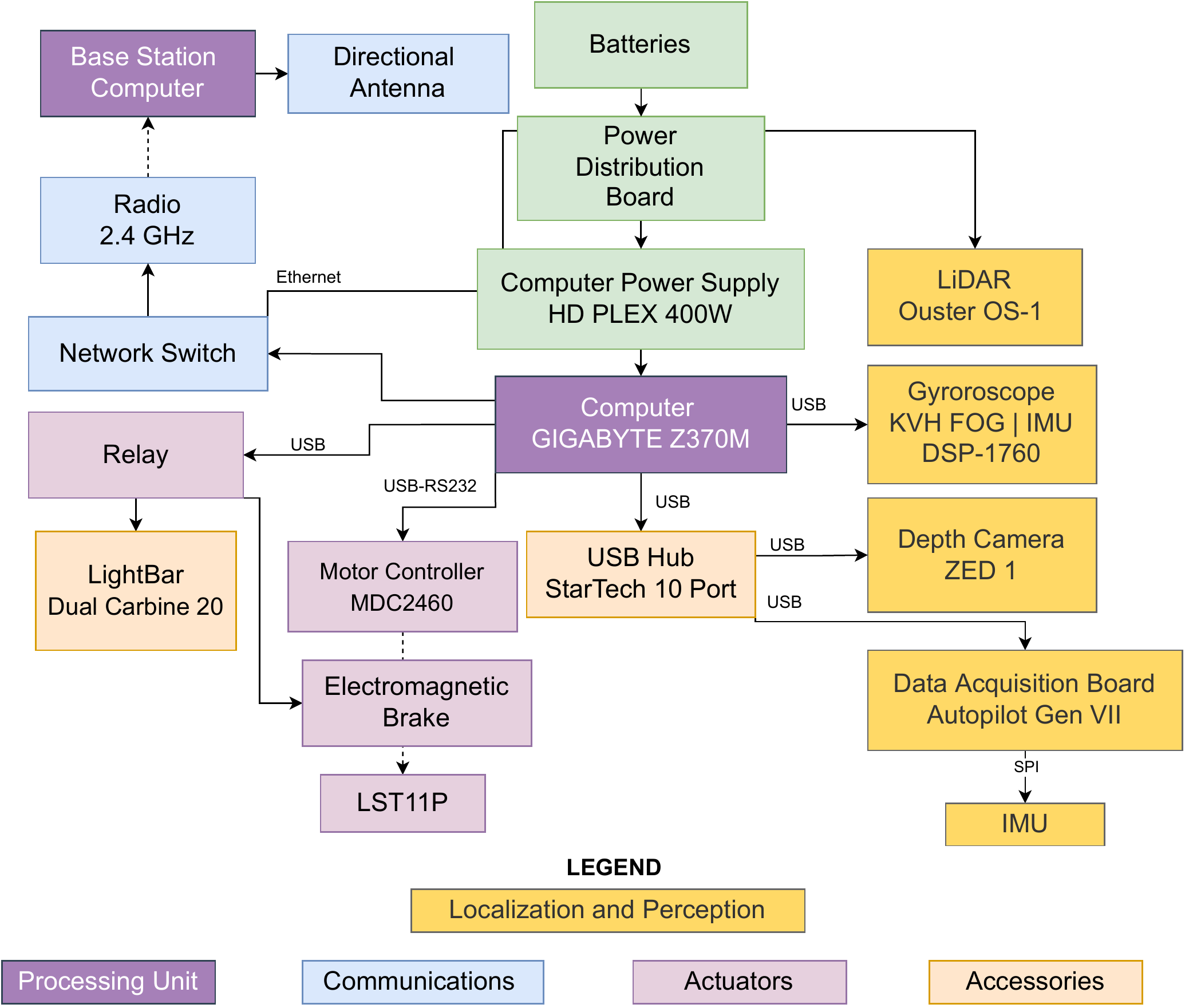}
\caption{Diagram showing Rhino's sensors and electronics required to complete the autonomous navigation and mapping tasks. 
\label{fig:electronics_diagram}}
\end{figure}

The Rhino robot is instrumented with a set of sensors to enable autonomous operation in underground environments. Figure \ref{fig:electronics_diagram} shows an overview of the rover's electronics and sensors. The ZED Stereo Camera and an Ouster OS1-64 LiDAR with 64 channels and a max range of 200m and $45 ^{\circ}$ vertical field of view are used to sense the environment. The Analog Devices ADIS16495 Inertial Measurement Unit (IMU) that has a triaxial digital gyroscope and a triaxial accelerometer and a KVH DSP-1760 single-axis fiber optic gyroscope are used in combination with the exteroceptive sensors to reduce errors in the localization and mapping estimates. 
The robot was designed so that IMU was mounted below the LiDAR with only a translation offset in the Z-direction.
The LiDAR is extrinsically calibrated with an IMU to track the motion distortion in each LiDAR scan \cite{jonasthesis}. Data is collected and processed onboard the robot using an Intel i7-9700K computer with 32GB of RAM memory and a GTX 1050Ti GPU.

Due to the high current demand for the motors and drone, and the sensitivity of the sensors and electronics such as the computer, the power distribution of the robot is split into three independent modules: the electronics and sensors, the drone, and the drive motors.
Four batteries are dedicated to powering the motors, which is an independent system due to the large current spikes and back-EMF. Two batteries are assembled in series ($\SI{25.6}{\V}$) to power the motors on each side of the robot. For modularity of the drone system, there are four batteries dedicated to the UAV system, since the UGV and UAV operate independently in most situations. This system uses one $\SI{12.8}{\V}$ battery for powering the winch motor, linear actuators and electronics and $\SI{38.4}{\V}$ to power the UAV. The rest of the electronics and sensors onboard are powered by the other two $\SI{12.8}{\V}$ batteries in series that are connected to a custom power distribution board that is able to supply a wide range of voltages from $\SI{3.3}{\V}$ to $\SI{48}{\V}$.

Rhino's drive motors are four LST11P 24 VDC brushed motors with a gear reduction of $45.5:1$ and electromagnetic brakes. Two 2-Channel Roboteq MDC2460 motor controllers are used to control the motors in a skid steer configuration and provide up to $\SI{50}{\A}$ maximum current and $\SI{10}{\A}$ continuous current per motor. An additional circuit was designed to protect the rest of the system from high voltage transients caused by the back-EMF due to the high inertia of Rhino. The circuit involves adding a relay, fuse, fly-back diode, an emergency-stop (e-stop) button, and an In-Rush current limiter to the motor's power module. The In-Rush Current limiter protects the motor controllers from the large impulse of current when the robot is switched on. The relay and e-stop act as a switch to power the motors on and off and also allows power to the motors to be cut off in emergency situations. On the software side of the driving configuration, the acceleration is linear and limited to $\SI{0.5}{\meter\per\s\squared}$ and the Proportional-Integral (PI) feedback controller has a limitation on the maximum integral value ensuring smoother driving. In order to control the brakes during Rhino's missions and for safety, the electromagnetic brakes are interfaced with a relay that allows user input. 

Rhino is instrumented with light bars that are controlled by USB relays for improved visibility during underground operations. The robot also has four LED indicators on its side that provide visual information on whether it is in autonomous or teleoperation mode and the status of the brakes.

The goal of Rhino's communication system is to enable the transmission of data between the robot and a base station computer when in direct line of sight. Data such as the orientation of the system, camera feedback during teleoperation mode, state of the brakes, the current draw of the motors, localization pose estimates, etc. 
The robot communication system consists of two omnidirectional antennas with a $\SI{2.4}{\GHz}$ and $\SI{900}{\MHz}$ frequency radio system by Ubiquiti interfaced with the system via a network switch.
A base station computer with high-gain directional antennas is used to communicate with the robot. The communication link transmits crucial information such as health status and sensor data; allows for joystick teleoperation from the base and for emergency braking; and is also used to deploy and monitor the robot on autonomous missions. This video shows the robot being fully assembled\footnote{\url{https://youtu.be/PvkYFZXjrGM}}.

\section{Robot Software and Operations}
\label{sec:robot_software_and_operations}

\begin{figure*}[b]
\centering
\includegraphics[width=0.85\textwidth]{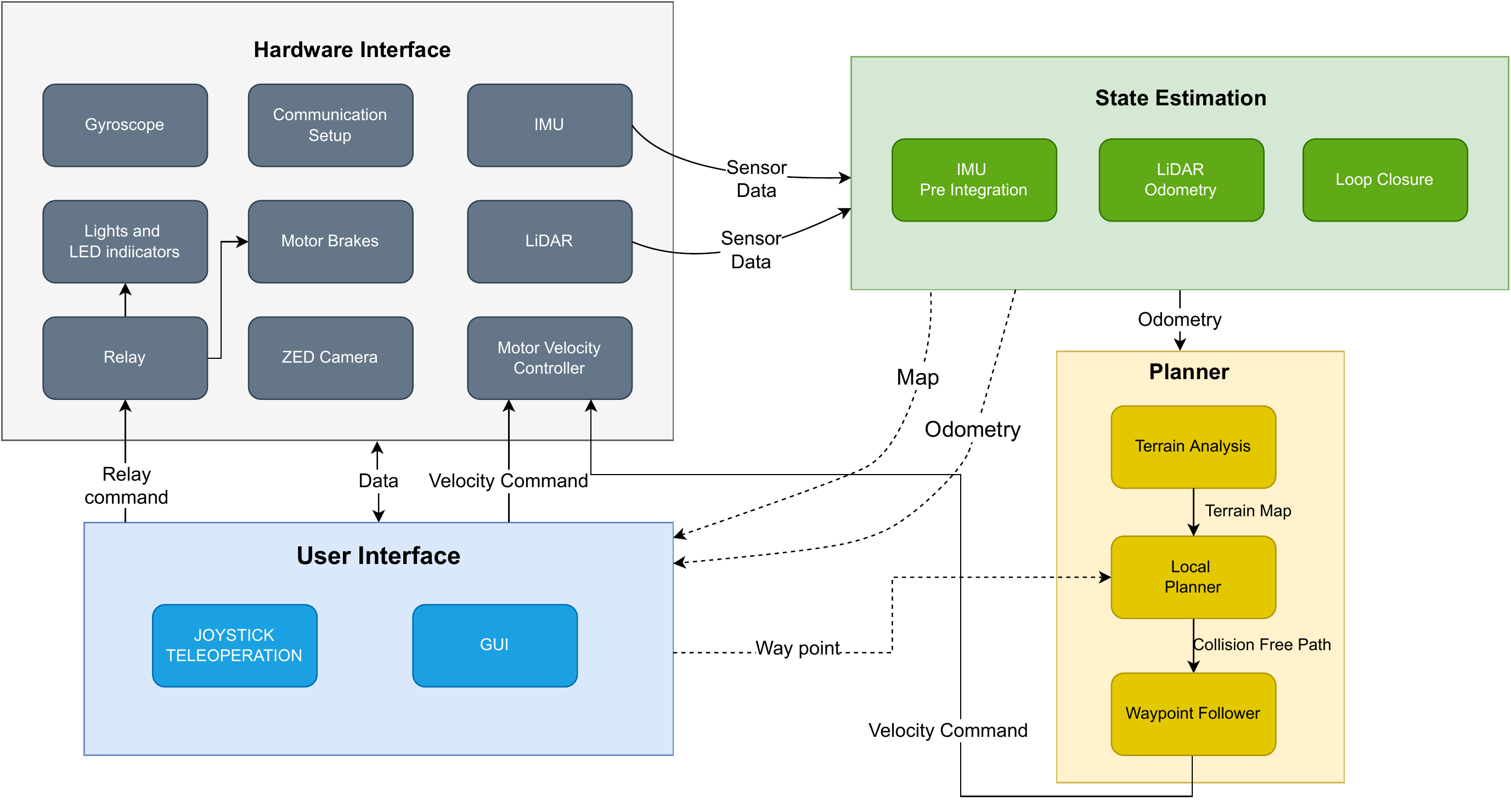}
\caption{Systems Software Architecture, where grey is the hardware interface module, green is the state estimation module, yellow is the planner module, and blue is the user interface module. \label{fig:overview}}
\end{figure*}

The goal of the robot operation is to enable the robot to drive using waypoint commands inside the mine environment while avoiding obstacles and maintaining a precise state estimation. These commands can be either sent by a user, a task planner algorithm like \cite{tatsch2020route} or an exploration-based algorithm like \cite{yamauchi1997frontier}, \cite{bircher2016receding}, \cite{cao2021tare}. The software is developed in ROS \cite{quigley2009ros} and all core components are running on the robot computer. The software architecture can be divided into four modules: Hardware Interface, State Estimation, Planning, and User Interface, as shown in Figure \ref{fig:overview}.

The Hardware interface module involves developing software that allows for communication with the sensors and electronics on Rhino which are essential for all the modules. The skid-steering velocity controller algorithm uses command velocities as input at a $\SI{50}{\Hz}$ frequency and outputs individual motor commands via a serial interface with the motor controller. This controller also implements a ramp for the velocity commands, limiting the robot acceleration to $\SI{0.5}{\meter\per\s\squared}$. The IMU and FOG gyro data are also read using the serial interface at $\SI{200}{\Hz}$ and $\SI{100}{\Hz}$ respectively. The Ouster LiDAR and the ZED camera data are acquired using the default software provided by the developers. The USB relay software was developed for the activation and deactivation of the spotlights, LED indicators, and electromagnetic brakes. The communication is set up with the robot as the router server and a base station as a client to that computer, allowing for continuous operation even when the connection with the base is lost.

The state estimation module is responsible for providing the robot with reliable localization during autonomous navigation tasks. All the goal commands to the robot are in a global reference frame, therefore having a good state estimation is necessary.
The tightly-coupled LiDAR-inertial-odometry via smoothing and mapping (LIO-SAM) SLAM framework \cite{shan2020lio} formulated as a Maximum a Prior (MAP) problem was used since it has demonstrated high accuracy and real-time performance in other environments with similar characteristics such as the tunnel environment in the DARPA subT competition. LIO-SAM is built on a factor graph formulation that allows multisensor input such as IMU, LiDAR, etc., and allows for global optimization. Prior to the IMU data being used by the state estimation module it is calibrated using the Allan Variance method \cite{el2007analysis} and a complementary filter \cite{valenti2015keeping} implemented to infer the orientation estimates. To mitigate issues with pointcloud deskewing due to high acceleration, a nonlinear motion model that receives IMU measurements is used to estimate the motion of the vehicle during the LiDAR scan. The estimated motion from the IMU is also used as an initial guess in the LiDAR odometry optimization and the approximation of IMU biases in the factor graph. For real-time performance and to improve computational complexity, the framework utilizes scan matching at the local scale, as well as the use of selective keyframes. With the formulation being a MAP problem, a Gaussian noise model can be assumed and solved as a linear least-squares problem. The factor graph has nodes that represent the robot state, while the factors in the graph are IMU preintegration, LiDAR odometry, GPS, and loop closure. The state estimation module provides odometry estimates at $\SI{10}{\Hz}$ and is also responsible for creating maps of the environment.

The Planning module is responsible for receiving $(x, y, z)$ waypoints commands and returning velocity commands to the robot. It includes a terrain analysis, local planner, and path-following modules. The terrain analysis subscribes to the state estimation odometry and LiDAR and map data and generates a $\SI{15}{\meter}$ by $\SI{15}{\meter}$ terrain map around the robot. It uses voxels \cite{oleynikova2017voxblox} to represent the terrain around the robot and estimates the ground height. Areas that are further than $\SI{0.2}{\meter}$ away from the ground are considered non-traversable. The map around the robot and the current LiDAR reading are merged to create an updated terrain map around the robot at every timestep at an update rate of $\SI{5}{\Hz}$. The terrain map and the waypoint goal are used as inputs for the local planner algorithm. The local planner algorithm uses FALCO \cite{zhang2020falco}, where motion primitives are pre-computed. The motion primitives around the robot are sampled via the Monte-Carlo method. A path is considered to be valid when every point within the robot's footprint is traversable.
The path with maximum likelihood towards the waypoint goal is chosen, while the local planner updates at a rate of $\SI{10}{\Hz}$. The path follower algorithm uses the odometry and the goal path to generate a velocity command that is sent to the motor controller at $\SI{20}{\Hz}$. A lookahead distance of $\SI{0.75}{\meter}$ is used to follow the selected path. A proportional-integral controller is used to define the linear $x$ and angular $z$ velocities at every timestep. The path follower algorithm is the only portion of the robot software where positions are converted and modeled with respect to the robot frame.

The user interface module was developed for ease of deployment of the robot in the underground mine environment. It consists of visualization tools and teleoperation. The user interface is designed to run on the base station computer, with communication routed through both the 900Mhz and the $\SI{2.4}{\GHz}$ antennas. A qt-based \cite{blanchette2006c++} graphical user interface was developed to provide the user with feedback on the robot's operation when the communication link is still available. This means monitoring the robot's health including velocity and the current being drawn by each motor, its temperature, battery level, and state estimates. Operators are able to monitor operations via the camera and point cloud, where the quality deteriorates with lower bandwidths. Waypoint goals can be sent by the user using the terminal, and any task planner or autonomous navigation algorithm. The teleoperation module is designed to have a higher priority than the velocity commands from the path follower in case of emergencies. The joystick can be used on both the robot and the base station computer, to send velocity commands to the velocity controller, and operate the spotlights. The electromagnetic brakes can only be released with the joystick by the user and it overwrites any motor command, therefore, increasing the safety of robot and operators while testing.

The developed robot software enables the robot to operate autonomously inside the subterranean environment. The state estimation module provides reliable odometry, allowing continuous mapping of the terrain around the robot and analysis of its traversability. This information is used to plan paths that the robot has to follow to reach its goals. The goals can be set either by the user via its user interface or an autonomous decision-making algorithm such as a task planner or an exploration algorithm.

\section{Experiments and Results}
\label{sec:experiments_and_results}


\begin{figure}
\centering
\includegraphics[width=0.95\linewidth]{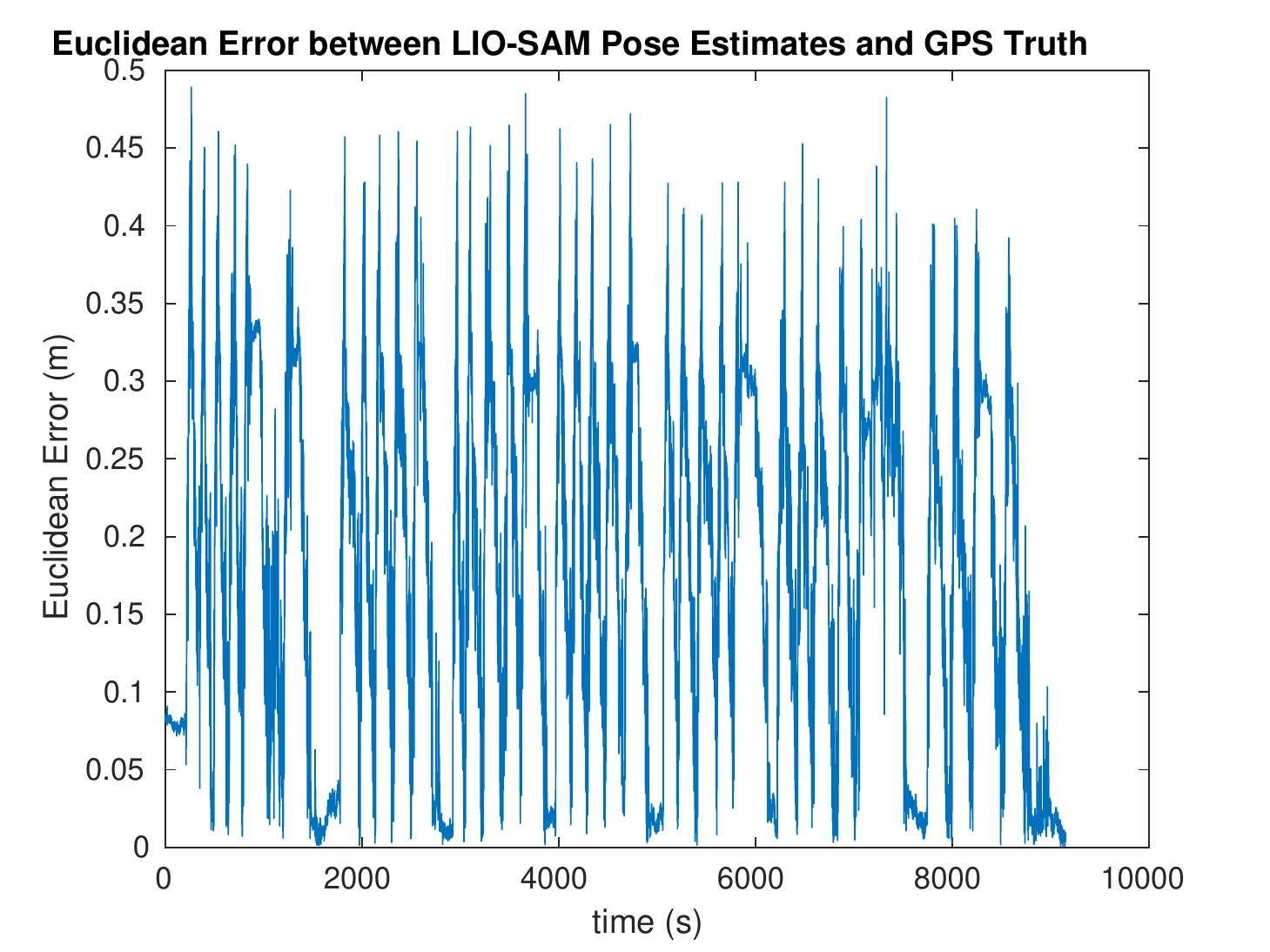}
\caption{Euclidean distance error vs time for the robot driving between 2 waypoints for $3.60 km$ \label{fig:back_and_forth_error}}
\end{figure}   
\unskip

Experiments were performed to validate and fine-tune the robot's capabilities. 
Initial tests were conducted to assess the robot's driving performance through teleoperation over different types of terrain including concrete, grass, tiled indoor surfaces, gravel, and muddy terrain. 
Localization, waypoint navigation, and obstacle avoidance capabilities were tested on several terrains with the most extensive tests being carried out on an outdoor gravel terrain, indoor obstacle course, and inside an underground mine. 

\subsection{Experiments on university campus}
The robot was initially tested on a large gravel terrain to verify the performance of the robot localization by using user-generated waypoints as inputs for the planner module to allow the system to autonomously navigate to multiple waypoints for an extended period of time and validated via a truth reference solution from GPS measurements.
The truth reference solution is determined by a carrier-phase differential GPS (DGPS) setup.
The setup for the DGPS solution consists of two dual-frequency Novatel OEM-615 GPS receivers, and L1/L2 Pinwheel antennas, one mounted to the rover and the other mounted on a base station \cite{kilic2021}. The GPS data is post-processed and is solely used to validate the state estimates.
The robot was deployed using the base station and monitored using the developed user interface during tests. In the first test, the robot autonomously navigated between waypoints which were set $\SI{100}{\meter}$ apart multiple times. The rover drove for $150$ minutes and a total distance of $\SI{3.6}{\km}$ in a single run. Figure \ref{fig:back_and_forth_error} shows the LIO-SAM euclidean distance error compared to the RTK GPS ground truth, where the error throughout the entire run is bounded within $\SI{0.5}{\meter}$ and a \SI{0.2}{\meter} root mean squared error.

A second test was indoors, where an obstacle course was created using card boxes, and different patterns that can be encountered on the mine were evaluated. The obstacle courses included narrow passages $\SI{2}{\meter}$ wide and maze-like patterns, where the robot has to zigzag through the obstacle walls and there are dead ends where the waypoint is not directly in front of the robot and it is required to reverse to a previous point before it can reach its goal. These tests evaluated the robot's obstacle detection, localization in a narrower environment, planning, and obstacle avoidance capabilities. The robot was able to reliably traverse through the obstacle course in all designed maps.

Besides validating localization, these tests also demonstrate the capability of the robot to operate for extended periods of time, being able to drive more than 6 hours with a single battery cycle.

\subsection{Underground Mine}

\begin{figure}
\centering
\includegraphics[width=0.95\linewidth]{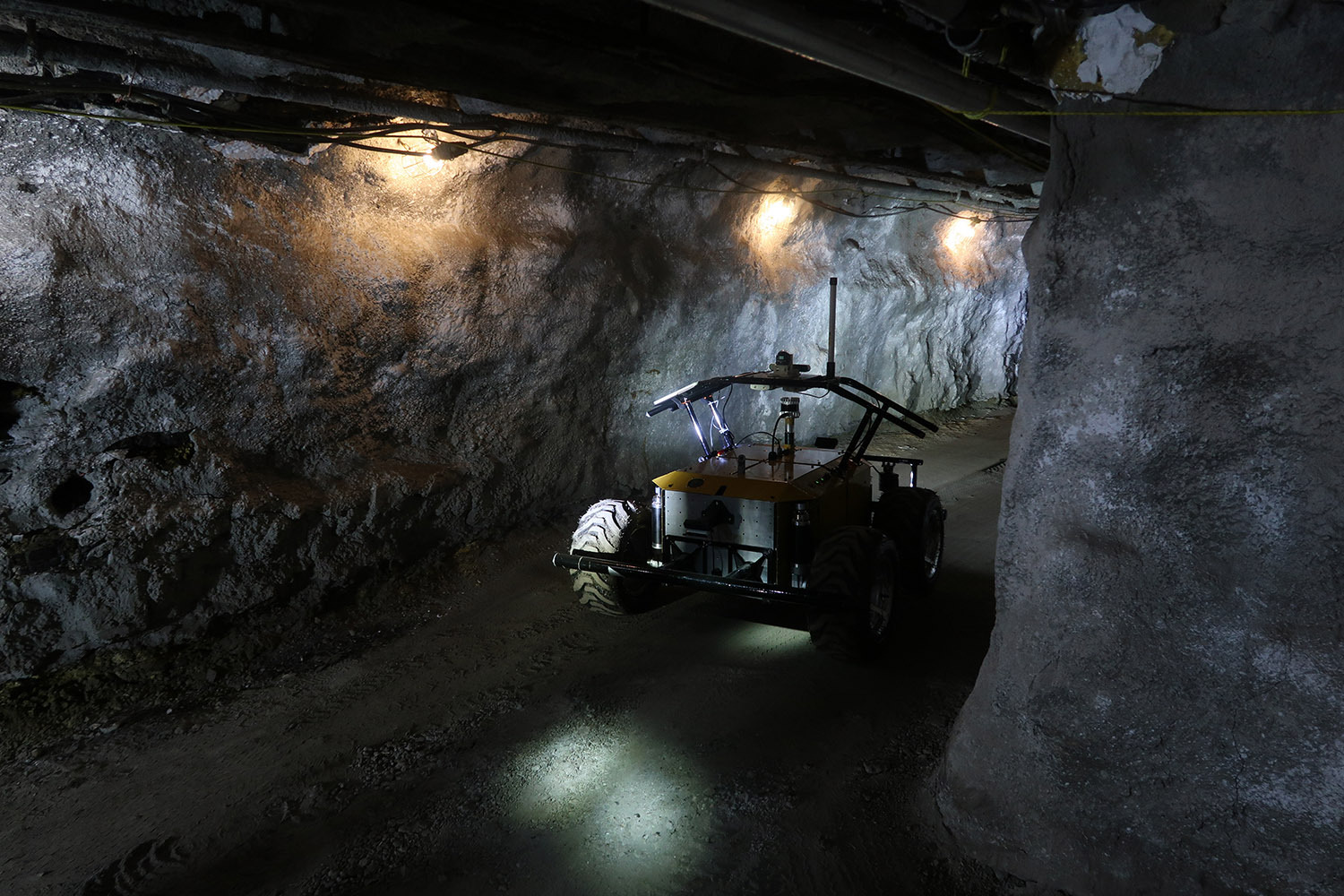}
\caption{Rhino Robot inside the experimental Mine while mapping. \label{fig:mine_env}}
\end{figure}

The robot's ability to operate in subterranean environments was validated on multiple visits to the National Institute for Occupational Safety and Health (NIOSH) Safety Research Coal Mine, Pittsburgh, Pennsylvania. Figure \ref{fig:mine_env} shows the robot inside the mine and a video of the robot operating inside the mine can be seen here\footnote{\url{https://www.youtube.com/watch?v=ouDQXcnTSxA&ab_channel=WVUIRL}}.

\begin{figure}[]
\centering
\includegraphics[width=0.9\linewidth]{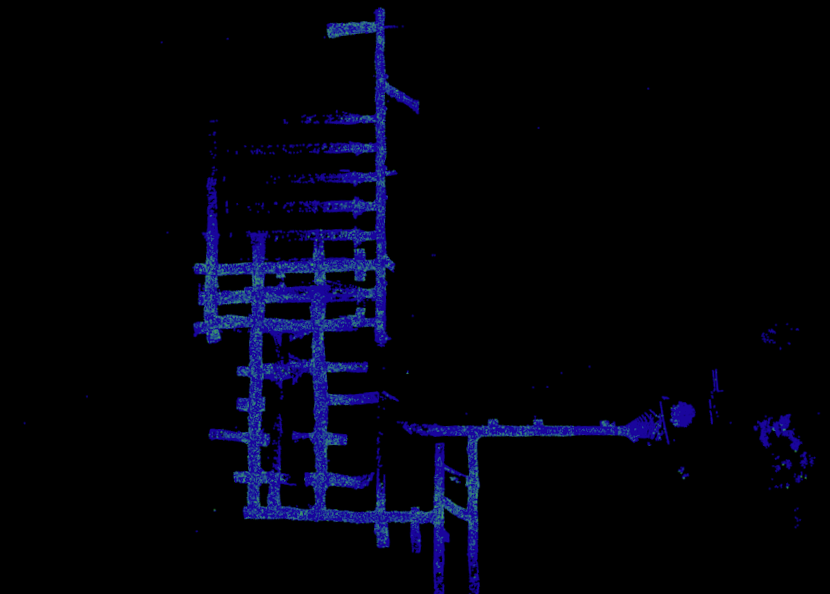}
\caption{Underground Mine LiDAR Mapping \label{fig:mine map}}
\end{figure}

In order to verify the system's ability to operate in the mine, driving and mapping tests were performed by setting pre-defined waypoint goals for the robot via the base station. Rhino was able to traverse the underground terrain while maintaining its state estimation always returning to a starting position and parking within $\SI{50}{\cm}$ of that position. Figure \ref{fig:mine map} shows the map built by the rover during a run where Rhino traversed $\SI{2.07}{\km}$ inside the mine. During this mission, the robot drove over different types of terrain such as gravel, dry, and muddy terrain and built a mine map using the localization and mapping module. During the mapping missions, the loop closure detection was qualitatively evaluated by driving in several loops to assess the system's ability to recognize loop closures. The areas of the mine that were not traversed had large amounts of water that the operators deemed unsafe for the system, although the Rhino was able to navigate through some sections with water.

The goal of the communications test was to verify the capabilities of the communications setup instrumented on Rhino. 
The operators were able to drive line-of-sight over the communications setup by using the user-interface.
Although the operators could drive Rhino and monitor the generation of the 3D map, some limitations were discovered.
The limitations were a result of the harsh mine environment attenuating the radio signals.
This was discovered when the robot was driven out-of-line-sight of the base station directional antenna, the operators would instantly lose connection to the system.

\begin{figure*}[t]
\centering
\includegraphics[width=0.95\linewidth, trim={0cm 1cm 0cm 4cm}, clip]{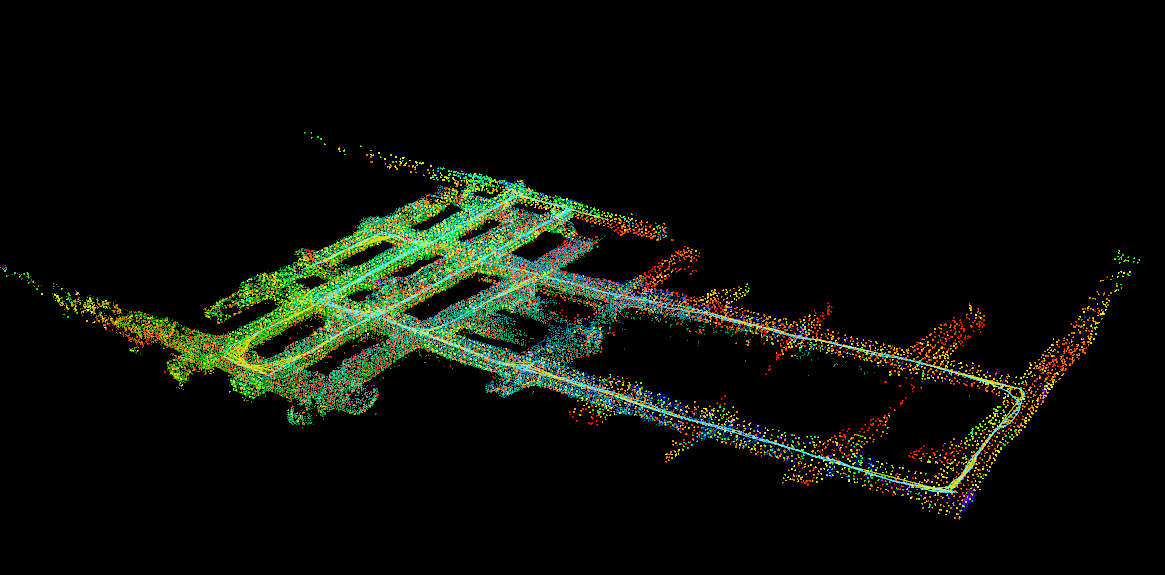}
\caption{Underground Mine Driven Path \label{fig:mine_path}}
\end{figure*}

For testing the robot's autonomous navigation system, a set of free valid waypoints are generated based on the map previously built using teleoperation. These waypoints are sent to the robot in sequence. The robot switches to waypoint driving mode without restarting the system allowing the system to utilize the localization solution previously generated. The mine map with the path driven by the robot is shown in Figure \ref{fig:mine_path}. The tests demonstrated Rhino's ability to follow the planned path autonomously and reach its goals, showing the repeatability of being able to drive the same $\SI{415}{\meter}$ loop five times while maintaining accurate localization and returning to the starting position every time.

A lesson learned during these tests is that a better understanding of the environment would enable the robot to drive more efficiently and have a safer autonomous driving mode. Examples of these situations are narrow passages where two-way driving is preferable compared to trying to maneuver following arcs to return to a position behind the robot. Knowing the terrain type, where turn-in-place maneuvers can be performed over surfaces with lower friction such as gravel and muddy terrain, but is more challenging over terrain with higher friction such as paved surfaces.

\section{Conclusions and Future Work}
\label{sec:conclusion}

\noindent This work presented Rhino, a robotic system for monitoring underground stone mines to improve safety and prevent future accidents. It is an unmanned ground vehicle equipped with various sensors and electronics that can navigate autonomously using a state-of-the-art mapping system. In practice, the system will be used to assess the structural integrity of the mine's roofs and pillars, allowing miners to take proactive measures to prevent accidents and evacuate if necessary. Here, the design was verified through testing outdoors and inside a mine environment. While Rhino was able to autonomously traverse difficult terrain for six continuous hours outdoors, it was able to accurately map the mine while avoiding obstacles and untraversable areas within. 

Future work includes more extensive tests in different types of mines and the development of new exploration algorithms to support fully autonomous operation. Another future work is the collaboration with the tethered drone, where the ground vehicle and the tethered drone can perform mapping maneuvers simultaneously, sharing its state estimation and mapping.

\section{Acknowledgments}
\noindent

This study was sponsored by the Alpha Foundation for the Improvement of Mine Safety and Health, Inc. (ALPHA FOUNDATION). The views, opinions, and recommendations expressed herein are solely those of the authors and do not imply any endorsement by the ALPHA FOUNDATION, its Directors and staff.

We would also like to thank Pittsburgh Mining Research Division (PMRD) employees (Charles Warren, Charlie Matthews, Mark Zuspan, Ted Klemetti, Brent Slaker, Nick Damiano, Nicole Evanek, and Mark Van Dyke) for facilitating experiments in NIOSH/PMRD Bruceton Experimental Mine and Safety Research Coal Mine

The researchers would like to express their gratitude to Tyler Wolf, Heath Cottrill, Matthew Collins and Henry Voss for their contributions to development of Rhino robot at the Interactive Robotics Lab at WVU. 
They would also like to thank Dr. Nicholas Ohi, Dr. Cagri Kilic, and Dr. Chizhao Yang for their support. 

We would also like to thank Dr. Guilherme Pereira, Bernardo Martinez, and Rogerio Lima, from the FARO Lab at WVU for the collaboration on the project and development of the tethered drone that is attached to the Rhino robot.

\bibliographystyle{apalike}
\bibliography{bibliography}

\end{document}